# Normalized online learning

**Stéphane Ross**
Carnegie Mellon University
Pittsburgh, PA, USA
stephaneross@cmu.edu

**Paul Mineiro**
Microsoft
Bellevue, WA, USA
paul.mineiro@gmail.com

**John Langford**
Microsoft Research
New York, NY, USA
jcl@microsoft.com

## Abstract

We introduce online learning algorithms which are independent of feature scales, proving regret bounds dependent on the ratio of scales existent in the data rather than the absolute scale. This has several useful effects: there is no need to pre-normalize data, the test-time and test-space complexity are reduced, and the algorithms are more robust.

## 1 Introduction

Any learning algorithm can be made invariant by initially transforming all data to a preferred coordinate system. In practice many algorithms begin by applying an affine transform to features so they are zero mean with standard deviation 1 [Li and Zhang, 1998]. For large data sets in the batch setting this preprocessing can be expensive, and in the online setting the analogous operation is unclear. Furthermore preprocessing is not applicable if the inputs to the algorithm are generated dynamically during learning, e.g., from an on-demand primal representation of a kernel [Sonnenburg and Franc, 2010], virtual example generated to enforce an invariant [Loosli et al., 2007], or machine learning reduction [Allwein et al., 2001].

When normalization techniques are too expensive or impossible we can either accept a loss of performance due to the use of misnormalized data or design learning algorithms which are inherently capable of dealing with unnormalized data. In the field of optimization, it is a settled matter that algorithms should operate independent of an individual dimensions scaling [Oren, 1974]. The same structure defines natural gradients [Wagenaar, 1998] where in the stochastic setting, results indicate that for the parametric case the Fisher metric is the unique invariant metric satisfying a certain regular and monotone property [Corcuera and Giummole, 1998]. Our interest here is in the online learning setting, where this structure is rare: typically regret bounds depend on the norm of features.

The biggest practical benefit of invariance to feature scaling is that learning algorithms "just work" in a more general sense. This is of significant importance in online learning settings where fiddling with hyper-parameters is often common, and this work can be regarded as an alternative to investigations of optimal hyper-parameter tuning [Bergstra and Bengio, 2012, Snoek et al., 2012, Hutter et al., 2013]. With a normalized update users do not need to know (or remember) to pre-normalize input datasets and the need to worry about hyper-parameter tuning is greatly reduced. In practical experience, it is common for those unfamiliar with machine learning to create and attempt to use datasets without proper normalization.

Eliminating the need to normalize data also reduces computational requirements at both training and test time. For particularly large datasets this can become important, since the computational cost in time and RAM of doing normalization can rival the cost and time of doing the machine learning (or even worse for naive centering of sparse data). Similarly, for applications which are constrained by testing time, knocking out the need for feature normalization allows more computational performance with the same features or better prediction performance when using the freed computational resources to use more features.

### 1.1 Adversarial Scaling

Adversarial analysis is fairly standard in online learning. However, an adversary capable of rescaling features can induce unbounded regret in common gradient descent methods. As an example consider the standard regret bound for projected online convex subgradient descent after $T$ rounds using the best learning rate in hindsight [Zinkevich, 2003],

$$R \leq \sqrt{T}||w^*||_2 \max_{t \in 1:T} ||g_t||_2.$$

Here $w^*$ is the best predictor in hindsight and $\{g_t\}$ is the sequence of instantaneous gradients encountered by the algorithm. Suppose $w^* = (1, 1) \in \mathbb{R}^2$ and imagine scaling the first coordinate by a factor of $s$. As $s \to \infty$, $||w^*||_2$ ap-proaches 1, but unfortunately for a linear predictor the gradient is proportional to the input, so $\max_{t\in 1:T} ||g_t||_2$ can be made arbitrarily large. Conversely as $s \to 0$, the gradient sequence remains bounded but $||w^*||_2$ becomes arbitrarily large. In both cases the regret bound can be made arbitrarily poor. This is a real effect rather than a mere artifact of analysis, as indicated by experiments with a synthetic two dimensional dataset in figure 1.1.

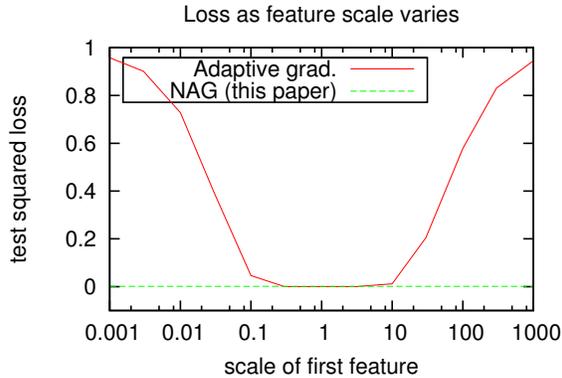

Figure 1: A comparison of performance of NAG (this paper) and adaptive gradient [McMahan and Streeter, 2010, Duchi et al., 2011] on a synthetic dataset with varying scale in the first feature.

Adaptive first-order online methods [McMahan and Streeter, 2010, Duchi et al., 2011] also have this vulnerability, despite adapting the geometry to the input sequence. Consider a variant of the adaptive gradient update (without projection)

$$w_{t+1} = w_t - \eta \operatorname{diag}(\sum_{s=1}^{t} g_s g_s^T)^{-1/2} g_t,$$

which has associated regret bound of order

$$||w^*||_2 \, d^{1/2} \sqrt{\inf_S \left\{ \sum_{t=1}^T \langle g_t, S^{-1} g_t \rangle : S \succeq 0, \operatorname{tr}(S) \leq d \right\}}.$$

Again by manipulating the scaling of a single axis this can be made arbitrarily poor.

The online Newton step [Hazan, 2006] algorithm has a regret bound independent of units as we address here. Unfortunately ONS space and time complexity grows quadratically with the length of the input sequence, but the existence of ONS motivates the search for computationally viable scale invariant online learning rules.

Similarly, the second order perceptron [Cesa-Bianchi et al., 2005] and AROW [Crammer et al., 2009] partially address this problem for hinge loss. These algorithms are not unit-free because they have hyperparameters whose optimal value varies with the scaling of features and again have running times that are superlinear in the dimensionality. More recently, diagonalized second order perceptron and AROW have been proposed [Orabona et al., 2012]. These algorithms are linear time, but their analysis is generally not unit free since it explicitly depends on the norm of the weight vector. Corollary 3 is unit invariant. A comparative analysis of empirical performance would be interesting to observe.

The use of unit invariant updates have been implicitly studied with asymptotic analysis and empirics. For example [Schaul et al., 2012] uses a per-parameter learning rate proportional to an estimate of gradient squared divided by variance and second derivative. Relative to this work, we prove finite regret bound guarantees for our algorithm.

### 1.2 Contributions

We define normalized online learning algorithms which are invariant to feature scaling, then show that these are interesting algorithms theoretically and experimentally.

We define a scaling adversary for online learning analysis. The critical additional property of this adversary is that algorithms with bounded regret must have updates which are invariant to feature scale. We prove that our algorithm has a small regret against this more stringent adversary.

We then experiment with this learning algorithm on a number of datasets. For pre-normalized datasets, we find that it makes little difference as expected, while for unnormalized or improperly normalized datasets this update rule offers large advantages over standard online update rules. All of our code is a part of the open source Vowpal Wabbit project [Ross et al., 2012].

## 2 Notation

Throughout this draft, the indices $i, j$ indicate elements of a vector, while the index $t, T$ or a particular number indicates time. A label $y$ is associated with some features $x$, and we are concerned with linear prediction $\sum_i w_i x_i$ resulting in some loss for which a gradient $g$ can be computed with respect to the weights. Other notation is introduced as it is defined.

## 3 The algorithm

We start with the simplest version of a scale invariant online learning algorithm.

NG (Normalized Gradient Descent) is presented in algorithm 1. NG adds scale invariance to online gradient descent, making it work for any scaling of features within the dataset.

Without $s, N$, this algorithm simplifies to standard stochas-

**Algorithm 1** NG(learning_rate $\eta_t$)

1. Initially $w_i = 0$, $s_i = 0$, $N = 0$
2. For each timestep $t$ observe example $(x, y)$
   (a) For each $i$, if $|x_i| > s_i$
      i. $w_i \leftarrow \frac{w_i s_i^2}{|x_i|^2}$
      ii. $s_i \leftarrow |x_i|$
   (b) $\hat{y} = \sum_i w_i x_i$
   (c) $N \leftarrow N + \sum_i \frac{x_i^2}{s_i^2}$
   (d) For each $i$,
      i. $w_i \leftarrow w_i - \eta_t \frac{t}{N} \frac{1}{s_i^2} \frac{\partial L(\hat{y}, y)}{\partial w_i}$

---

**Algorithm 2** NAG(learning_rate $\eta$)

1. Initially $w_i = 0$, $s_i = 0$, $G_i = 0$, $N = 0$
2. For each timestep $t$ observe example $(x, y)$
   (a) For each $i$, if $|x_i| > s_i$
      i. $w_i \leftarrow \frac{w_i s_i}{|x_i|}$
      ii. $s_i \leftarrow |x_i|$
   (b) $\hat{y} = \sum_i w_i x_i$
   (c) $N \leftarrow N + \sum_i \frac{x_i^2}{s_i^2}$
   (d) For each $i$,
      i. $G_i \leftarrow G_i + \left(\frac{\partial L(\hat{y}, y)}{\partial w_i}\right)^2$
      ii. $w_i \leftarrow w_i - \eta \sqrt{\frac{t}{N}} \frac{1}{s_i \sqrt{G_i}} \frac{\partial L(\hat{y}, y)}{\partial w_i}$

---

tic gradient descent.

The vector element $s_i$ stores the magnitude of feature $i$ according to $s_{ti} = \max_{t' \in \{1...t\}} |x_{t'i}|$. These are updated and maintained online in steps 2.(a).ii, and used to rescale the update on a per-feature basis in step 2.(d).i.

Using $N$ makes the learning rate (rather than feature scale) control the average change in prediction from an update. Here $N/t$ is the average change in the prediction excluding $\eta$, so multiplying by $1/(N/t) = t/N$ causes the average change in the prediction to be entirely controlled by $\eta$.

Step 2.(a).i squashes a weight $i$ when a new scale is encountered. Neglecting the impact of $N$, the new value is precisely equal to what the weight's value would have been if all previous updates used the new scale.

Many other online learning algorithms can be made scale invariant using variants of this approach. One attractive choice is adaptive gradient descent [McMahan and Streeter, 2010, Duchi et al., 2011] since this also has per-feature learning rates. The normalized version of adaptive gradient descent is given in algorithm 2.

In order to use this, the algorithm must maintain the sum of gradients squared $G_i = \sum_{(x,y) \text{ observed}} \left(\frac{\partial L(\hat{y},y)}{\partial w_i}\right)^2$ for feature $i$ in step 2.d.i. The interaction between $N$ and $G$ is somewhat tricky, because a large average update (i.e. most features have a magnitude near their scale) increases the value of $G_i$ as well as $N$ implying the power on $N$ must be decreased to compensate. Similarly, we reduce the power on $s_i$ and $|x_i|$ to 1 throughout. The more complex update rule is scale invariant and the dependence on $N$ introduces an automatic global rescaling of the update rule.

In the next sections we analyze and justify this algorithm. We demonstrate that NAG competes well against a set of predictors $w$ with predictions ($w^\top x$) bounded by some constant over all the inputs $x_t$ seen during training. In practice, as this is potentially sensitive to outliers, we also consider a squared norm version of NAG, which we refer to as sNAG that is a straightforward modification—we simply keep the accumulator $s_i = \sum x_i^2$ and use $\sqrt{s_i/t}$ in the update rule. That is, normalization is carried using the standard deviation (more precisely, the square root of the second moment) of each feature, rather than the max norm. With respect to our analysis below, this simple modification can be interpreted as changing slightly the set of predictors we compete against, i.e. predictors with predictions bounded by a constant only over the inputs within 1 standard deviation. Intuitively, this is more robust and appropriate in the presence of outliers. While our analysis focuses on NAG, in practice, sNAG sometimes yield improved performance.

## 4 The Scaling Adversary Setting

In common machine learning practice, the choice of units for any particular feature is arbitrary. For example, when estimating the value of a house, the land associated with a house may be encoded either in acres or square feet. To model this effect, we propose a scaling adversary, which is more powerful than the standard adversary in adversarial online learning settings.

The setting for analysis is similar to adversarial online linear learning, with the primary difference in the goal. The setting proceeds in the following round-by-round fashion where

1. Prior to all rounds, the adversary commits to a fixed positive-definite matrix $S$. This is not revealed to the learner.

2. On each round $t$,
   (a) The adversary chooses a vector $x_t$ such that $||S^{1/2} x_t||_\infty \leq 1$, where $S^{1/2}$ is the principal

square root.
(b) The learner makes a prediction $\hat{y}_t = w_t^\top x_t$.
(c) The correct label $y_t$ is revealed and a loss $\ell(\hat{y}_t, y_t)$ is incurred.
(d) The learner updates the predictor to $w_{t+1}$.

For example, in a regression setting, $\ell$ could be the squared loss $\ell(\hat{y}, y) = (\hat{y} - y)^2$, or in a binary classification setting, $\ell$ could be the hinge loss $\ell(\hat{y}, y) = \max(0, 1 - y\hat{y})$. We consider general cases where the loss is only a function of $\hat{y}$ (i.e. no direct penalty on $w$) and convex in $\hat{y}$ (therefore convex in $w$).

Although step 1 above is phrased in terms of an adversary, in practice what is being modeled is "the data set was prepared using arbitrary units for each feature."

Step 2 (a) above is phrased in terms of $\infty$-norm for ease of exposition, but more generally can be considered any $p$-norm. Additionally, this step can be generalized to impose a different constraint on the inputs. For instance, instead requiring all points lie inside some $p$-norm ball, we could require that the second moment of the inputs, under some scaling matrix $S$ is 1. This is the model of the adversary for sNAG.

### 4.1 Competing against a Bounded Output Predictor

Our goal is to compete against the set of weight vectors whose output is bounded by some constant $C$ over the set of inputs the adversary can choose. Given step 2 (a) above, this is equivalent to $\mathcal{W}_X^C = \{w \mid ||S^{-1/2}w||_1 \leq C\}$, i.e., the set of $w$ with dual norm less than $C$. In other words, the regret $R_T$ at timestep $T$ is given by:

$$R_T = \sum_{t=1}^T \ell(\hat{y}_t, y_t) - \min_{w|\forall t, w^\top x_t \leq C} \sum_{t=1}^T \ell(w^\top x_t, y_t)$$

Here we use the fact that $\{w^\top x_t \leq C\} = \{w \mid ||S^{-1/2}w||_1 \leq C\}$. In the more general case of a $p$-norm for step 2 (a), we would choose $\mathcal{W}_X^C = \{w \mid ||S^{-1/2}w||_q \leq C\}$ for $q$ such that $\frac{1}{p} + \frac{1}{q} = 1$. Note that the "true" $S$ of the adversary is an abstraction. It is unknown and only partially revealed through the data. In our analysis, we will instead be interested to bound regret against bounded output predictors for an empirical estimate of $S$, defined by the minimum volume $L_p$ ball containing all observed inputs. For $p = \infty$, $\mathcal{W}_X^C$ for the "true" $S$ is always a subset of the predictors allowed under this empirical $S$ (assuming both are diagonal matrices). In general, this does not necessarily hold for all $p$ norms, but the empirical $S$ always allows a larger volume of predictors than the "true" $S$.

## 5 Analysis

In this section, we analyze scale invariant update rules in several ways. The analysis is structurally similar to that used for adaptive gradient descent [McMahan and Streeter, 2010, Duchi et al., 2011] with necessary differences to achieve scale invariance. We analyze the best solution in hindsight, the best solution in a transductive setting, and the best solution in an online setting. These settings are each a bit more difficult than the previous, and in each we prove regret bounds which are invariant to feature scales.

We consider algorithms updating according to $w_{t+1} = w_t - A_t^{-1} g_t$, where $g_t$ is the gradient of the loss at time $t$ w.r.t. $w$ at $w_t$, and $A_t$ is a sequence of $d \times d$ symmetric positive (semi-)definite matrices that our algorithm can choose. Both algorithms 1 and 2 fit this general framework. Combining the convexity of the loss function and the definition of the update rule yields the following result.

**Lemma 1.**

$$2R_T \leq (w_1 - w^*)A_1(w_1 - w^*)$$
$$+ \sum_{t=1}^T (w_t - w^*)^\top (A_{t+1} - A_t)(w_t - w^*)$$
$$+ \sum_{t=1}^T g_t^\top A_t^{-1} g_t.$$

*We defer all proofs to the appendix.*

### 5.1 Best Choice of Conditioner in Hindsight

Suppose we start from $w_1 = 0$ and before the start of the algorithm, we would try to guess what is the best fixed matrix $A$, so that $A_t = A$ for all $t$. In order to minimize regret, what would the best guess be? This was initially analyzed for adaptive gradient descent [McMahan and Streeter, 2010, Duchi et al., 2011]. Consider the case where $A$ is a diagonal matrix.

Using lemma 1, for a fixed diagonal matrix $A$ and with $w_1 = 0$, the regret bound is:

$$R_T \leq \frac{1}{2} \sum_{i=1}^d \left( A_{ii}(w_i^*)^2 + \sum_{t=1}^T \frac{g_{ti}^2}{A_{ii}} \right).$$

Taking the derivative w.r.t. $A_{ii}$, we obtain:

$$\frac{\partial}{\partial A_{ii}} \frac{1}{2} \sum_{i=1}^d \left( A_{ii}(w_i^*)^2 + \sum_{t=1}^T \frac{g_{ti}^2}{A_{ii}} \right)$$
$$= \frac{1}{2} \left( (w_i^*)^2 - \sum_{t=1}^T \frac{g_{ti}^2}{A_{ii}^2} \right).$$

Solving for when this is 0, we obtain

$$A_{ii}^* = \frac{\sqrt{\sum_{t=1}^T g_{ti}^2}}{|w_i^*|}.$$

For this particular choice of $A$, then the regret is bounded by

$$R_T \leq \sum_{i=1}^{d} \left( |w_i^*| \sqrt{\sum_{t=1}^{T} g_{ti}^2} \right).$$

We can observe that this regret is the same no matter the scaling of the inputs. For instance if any axis $i$ is scaled by a factor $s_i$, then $w_i^*$ would be a factor $1/s_i$ smaller, and the gradient $g_{ti}$ a factor $s_i$ larger, which would cancel out. Hence this regret can be thought as the regret the algorithm would obtain when all features have the same unit scale.

However, because of the dependency of $A$ on $w^*$, this does not give us a good way to approximate this with data we have observed so far. To remove this dependency, we can analyze for the best $A$ when assuming a worst case for $w^*$. This is the point at which the analysis here differs from adaptive gradient descent where the dependence on $w^*$ was dropped.

**Lemma 2.** *Let $S$ be the diagonal matrix with minimum determinant (volume) s.t. $||S^{1/2}x_i||_p \leq 1$ for all $i \in 1:T$. The solution to*

$$\min_{A} \max_{w^* \in \mathcal{W}_X^C} \frac{1}{2} \sum_{i=1}^{d} \left( A_{ii}(w_i^*)^2 + \sum_{t=1}^{T} \frac{g_{ti}^2}{A_{ii}} \right)$$

*is given by*

$$A_{ii}^* = \frac{1}{C} \sqrt{\frac{\sum_{t=1}^{T} g_{ti}^2}{S_{ii}}},$$

*and the regret bound for this particular choice of $A$ is given by*

$$R_T \leq C \sum_{i=1}^{d} \sqrt{S_{ii} \sum_{t=1}^{T} g_{ti}^2}.$$

Again the value of the regret bound does not change if the features are rescaled. This is most easily appreciated by considering a specific norm. The simplest case is for $p = \infty$ where the coefficients $S_{ii}$ can be defined directly in terms of the range of each feature, i.e. $S_{ii} = \frac{1}{\max_t |x_{ti}|^2}$. Thus for $p = \infty$, we can choose

$$A_{ii}^* = \frac{1}{C} \sqrt{\sum_{t=1}^{T} g_{ti}^2} \max_{t \in 1:T} |x_{ti}|,$$

leading to a regret of

$$R_T \leq C \sum_{i=1}^{d} \frac{\sqrt{\sum_{t=1}^{T} g_{ti}^2}}{\max_{t \in 1:T} |x_{ti}|}.$$

The scale invariance of the regret bound is now readily apparent. This regret can potentially be order $O(d\sqrt{T})$.

### 5.2 $p = 2$ case

For $p = 2$, computing the coefficients $S_{ii}$ is more complicated, but if you have access to the actual coefficients $S_{ii}$, the regret is order $O(\sqrt{dT})$. This can be seen as follows. Let $g_t' = \left.\frac{\partial \ell}{\partial \hat{y}}\right|_{\hat{y}_t, y_t}$ the derivative of the loss at time $t$ evaluated at the predicted $\hat{y}_t$. Then $g_{ti} = g_t' x_{ti}$ and we can see that:

$$\begin{aligned}
&\sum_{i=1}^{d} \sqrt{S_{ii} \sum_{t=1}^{T} g_{ti}^2} \\
&= d \sum_{i=1}^{d} \frac{1}{d} \sqrt{S_{ii} \sum_{t=1}^{T} g_{ti}^2} \\
&\leq d \sqrt{\sum_{i=1}^{d} \frac{1}{d} S_{ii} \sum_{t=1}^{T} g_{ti}^2} \\
&= \sqrt{d} \sqrt{\sum_{i=1}^{d} S_{ii} \sum_{t=1}^{T} g_{ti}^2} \\
&= \sqrt{d} \sqrt{\sum_{t=1}^{T} g_t'^2 \sum_{i=1}^{d} S_{ii} x_{ti}^2} \\
&\leq \sqrt{d} \sqrt{\sum_{t=1}^{T} g_t'^2}
\end{aligned}$$

where the last inequality holds by assumption. For $p = 2$, we have $R_T \leq C\sqrt{d}\sqrt{\sum_{t=1}^{T} g_t'^2}$.

### 5.3 Adaptive Conditioner

Lemma 2 does not lead to a practical algorithm, since at time $t$, we only observed $g_{1:t}$ and $x_{1:t}$, when performing the update for $w_{t+1}$. Hence we would not be able to compute this optimal conditioner $A^*$. However it suggests that we could potentially approximate this ideal choice using the information observed so far, e.g.,

$$A_{t,ii} = \frac{1}{C} \sqrt{\frac{\sum_{s=1}^{t} g_{si}^2}{S_{ii}^{(t)}}}, \qquad (1)$$

where $S^{(t)}$ is the diagonal matrix with minimum determinant s.t. $||S^{1/2}x_i||_p \leq 1$ for all $i \in 1:t$. There are two potential sources of additional regret in the above choice, one from truncating the sum of gradients, and the other from estimating the enclosing volume online.

#### 5.3.1 Transductive Case

To demonstrate that truncating the sum of gradients has only a modest impact on regret we first consider the transductive case, i.e., we assume we have access to all inputs $x_{1:T}$ that are coming in advance. However at time $t$, we do not know the future gradients $g_{t+1:T}$. Hence for this setting we could consider a 2-pass algorithm. On the first pass, compute the diagonal matrix $S$, and then on the second pass, perform adaptive gradient descent with the following conditioner at time $t$:

$$A_{t,ii} = \frac{1}{C\eta} \sqrt{\frac{\sum_{j=1}^{t} g_{ji}^2}{S_{ii}}}. \qquad (2)$$

We would like to be able to show that if we adapt the conditioner in this way, than our regret is not much worse than with the best conditioner in hindsight. To do so, we must introduce a projection step into the algorithm. The projection step enables us to bound the terms in lemma 1 corresponding to the use of a non-constant conditioner, which are related to the maximum distance between an intermediate weight vector and the optimal weight vector.

Define the projection $\Pi^A_{S,C,q}$ as

$$\Pi^A_{S,C,q}(w') = \underset{w \in \mathbb{R}^d \mid \|S^{-1/2}w\|_q \leq C}{\arg\min} (w-w')^\top A(w-w').$$

Utilizing this projection step in the update we can show the following.

**Theorem 1.** *Let $S$ be the diagonal matrix with minimum determinant s.t. $\|S^{1/2}x_i\|_p \leq 1$ for all $i \in 1:T$, and let $1/q = 1 - 1/p$. If we choose $A_t$ as in Equation 2 with $\eta = \sqrt{2}$ and use projection $w_{t+1} = \Pi^{A_t}_{S,C,q}(w_t - A_t^{-1}g_t)$ at each step, the regret is bounded by*

$$R_T \leq 2C\sqrt{2} \sum_{i=1}^d \sqrt{S_{ii} \sum_{j=1}^T g_{ji}^2}.$$

We note that this is only a factor $2\sqrt{2}$ worse than when using the best fixed $A$ in hindsight, knowing all gradients in advance.

### 5.3.2 Streaming Case

In this section we focus on the case $p = \infty$.

The transductive analysis indicates that using a partial sum of gradients does not meaningfully degrade the regret bound. We now investigate the impact of estimating the enclosing ellipsoid with a diagonal matrix online using only observed inputs,

$$A_{t,ii} = \frac{1}{C\eta} \sqrt{\sum_{j=1}^t g_{ji}^2 \max_{j \in 1:t} |x_{ji}|}. \quad (3)$$

The diagonal approximation is necessary for computational efficiency in NAG.

Intuitively the worst case is when the conditioner in equation 3 differs substantially from the transductive conditioner of equation 2 over most of the sequence. This is reflected in the regret bound below which is driven by the ratio between the first non-zero value of an input $x_{ji}$ encountered in the sequence and the maximum value it obtains over the sequence.

**Theorem 2.** *Let $p = \infty$, $q = 1$, and let $S^{(t)}$ be the diagonal matrix with minimum determinant s.t. $\|S^{1/2}x_i\|_p \leq 1$ for all $i \in 1 : t$. Let $\Delta_i = \frac{\max_{t \in 1:T} |x_{ti}|}{|x_{t_0^i i}|}$, for $t_0^i$ the first timestep the $i^{th}$ feature is non-zero. If we choose $A_t$ as in Equation 3, $\eta = \sqrt{2}$ and use projection $w_{t+1} = \Pi^{A_t}_{S^{(t)},C,q}(w_t - A_t^{-1}g_t)$ at each step, the regret is bounded by*

$$R_T \leq C \sum_{i=1}^d \frac{\sqrt{\sum_{j=1}^T g_{ji}^2}}{\max_{j \in 1:T} |x_{ji}|} \left( \frac{1 + 6\Delta_i + \Delta_i^2}{2\sqrt{2}} \right).$$

Comparing theorem 2 with theorem 1 reveals the degradation in regret due to online estimation of the enclosing ellipsoid. Although an adversary can in general manipulate this to cause large regret, there are nontrivial cases for which theorem 2 provides interesting protection. For example, if the non-zero feature values for dimension $i$ range over $[s_i, 2s_i]$ for some unknown $s_i$, then $1 \leq \Delta_i \leq 2$ and the regret bound is only a constant factor worse than the best choice of conditioner in hindsight.

Because the worst case streaming scenario is when the initial sequence has much lower scale than the entire sequence, we can improve the bound if we weaken the ability of the adversary to choose the sequence order. In particular, we allow the adversary to choose the sequence $\{x_t, y_t\}_{t=1}^T$ but then we subject the sequence to a random permutation before processing it. We can show that with high probability we must observe a high percentile value of each feature after only a few datapoints, which leads to the following corollary to theorem 2.

**Corollary 1.** *Let $\{x_t, y_t\}_{t=1}^T$ be an exchangeable sequence with $x_t \in \mathbb{R}^d$. Let $p = \infty$, $q = 1$, and let $S^{(t)}$ be the diagonal matrix with minimum determinant s.t. $\|S^{1/2}x_i\|_p \leq 1$ for all $i \in 1:t$. Choose $\delta > 0$ and $\nu \in (0,1)$. Let $\Delta_i = \frac{\max_{t \in 1:T} |x_{ti}|}{\max_{t \in 1:\tau} |x_{ti}|}$, where*

$$\tau = \left\lceil \frac{\log(d/\delta)}{\nu} \right\rceil.$$

*If $R_{\max}$ is the maximum regret that can be incurred on a single example, then choosing $\eta = \sqrt{2}$, and using projection $w_{t+1} = \Pi^{A_t}_{S^{(t)},C,q}(w_t - A_t^{-1}g_t)$ at each step, the regret is bounded by*

$$R_T \leq \left\lfloor \frac{\log(d/\delta)}{\nu} \right\rfloor R_{\max}$$
$$+ C \sum_{i=1}^d \frac{\sqrt{\sum_{j=1}^T g_{ji}^2}}{\max_{j \in 1:T} |x_{ji}|} \left( \frac{1 + 6\Delta_i + \Delta_i^2}{2\sqrt{2}} \right),$$

*and with probability at least $(1-\delta)$ over sequence realizations, for all $i \in 1:d$,*

$$\Delta_i \leq \frac{\max_{t \in 1:T} |x_{ti}|}{\text{Quantile}(\{|x_{ti}|\}_{t=1}^T, 1-\nu)},$$

*where Quantile$(\cdot, 1-\nu)$ is the $(1-\nu)$-th quantile of a given sequence.*

The quantity $R_{\max}$ can be related to $C$, if when making predictions, we always truncate $w_t^\top x_t$ in the interval $[-C, C]$. For instance, for the hinge loss and logistic loss, $R_{\max} \leq C+1$ if we truncate our predictions this way. Similarly for the squared loss, $R_{\max} \leq 4C \max(C, \max_t |y_t|)$. Although in theory an adversary can manipulate the ratio between the maximum and an extreme quantile to induce arbitrarily bad regret (i.e. make $\frac{\max_{t \in 1:T} |x_{ti}|}{\text{Quantile}(\{|x_{ti}|\}_{t=1}^T, 1-\nu)}$ arbitrarily large even for small $\nu$), in practice we can often expect this quantity to be close to $1$[1], and thus corollary 1 suggests that we may perform not much worse than when the scale of the features are known in advance. Our experiments demonstrate that this is the common behavior of the algorithm in practice.

| Dataset | Size | Features | Scale Range |
|---|---|---|---|
| Bank | 45,212 | 7 | [ 31, 102127 ] |
| Census | 199,523 | 13 | [ 2, 99999 ] |
| Covertype | 581,012 | 54 | [ 1, 7173 ] |
| CT Slice | 53,500 | 360 | [ 0.158, 1 ] |
| MSD | 463,715 | 90 | [ 60, 65735 ] |
| Shuttle | 43,500 | 9 | [ 105, 13839 ] |

Table 1: Datasets used for experiments. CT Slice and MSD are regression tasks, all others are classification. The scale of a feature is defined as the maximum empirical absolute value, and the scale range of a dataset defined as the minimum and maximum feature scales.

## 6 Experiments

Table 2 compares a variant of the normalized learning rule to the adaptive gradient method [McMahan and Streeter, 2010, Duchi et al., 2011] with $p = 2$ and without projection step for both algorithms. For each data set we exhaustively searched the space of learning rates to optimize average progressive validation loss. Besides the learning rate, the learning rule was the only parameter adjusted between the two conditions. The loss function used depended upon the task associated with the dataset, which was either 0-1 loss for classification tasks or squared loss for regression tasks. For regression tasks, the loss is divided by the worst possible squared loss, i.e., $(\max - \min)^2$.

The datasets utilized are: *Bank*, the UCI [Frank and Asuncion, 2010] Bank Marketing Data Set [Moro et al., 2011]; *Census*, the UCI Census-Income KDD Data Set; *Covertype*, the UCI Covertype Data Set; *CT Slice*, the UCI Relative Location of CT Slices on Axial Axis Data Set; *MSD*, the Million Song Database [Bertin-Mahieux et al., 2011]; and *Shuttle*, the UCI Statlog Shuttle Data Set. These were selected as public datasets plausibly exhibiting varying scales or lack of normalization. On other pre-normalized datasets which are publicly available, we observed relatively little difference between these update rules. To demonstrate the effect of pre-normalization on these data sets, we constructed a pre-normalized version of each one by dividing every feature by its maximum empirical absolute value.

[1]For instance, if $|x_{ti}|$ are exponentially distributed, $\Delta_i$ is roughly less than $\log(T/\delta)/\log(1/\nu)$ with probability at least $1 - \delta$, thus choosing $\nu = T^{-\alpha}$, for $\alpha \in (0, 0.5]$ makes this a small constant of order $\alpha^{-1} \log(1/\delta)$, while keeping the first term involving $R_{\max}$ order of $T^\alpha \leq \sqrt{T}$.

| Dataset | NAG | | AG | |
|---|---|---|---|---|
| | $\eta^*$ | Loss | $\eta^*$ | Loss |
| Bank | 0.55 | **0.098** | $5.5 * 10^{-5}$ | 0.109 |
| (Maxnorm) | 0.55 | 0.099 | 0.061 | 0.099 |
| Census | 0.2 | **0.050** | $1.2 * 10^{-6}$ | 0.054 |
| (Maxnorm) | 0.25 | 0.050 | $8.3 * 10^{-3}$ | 0.051 |
| Covertype | 1.5 | **0.27** | $5.6 * 10^{-7}$ | 0.32 |
| (Maxnorm) | 1.5 | 0.27 | 0.2 | 0.27 |
| CT Slice | 2.7 | 0.0023 | 0.022 | 0.0023 |
| (Maxnorm) | 2.7 | 0.0023 | 0.022 | 0.0023 |
| MSD | 9.0 | **0.0110** | $5.5 * 10^{-7}$ | 0.0130 |
| (Maxnorm) | 9.0 | 0.0110 | 6.0 | 0.0108 |
| Shuttle | 7.4 | 0.036 | $7.5 * 10^{-4}$ | 0.040 |
| (Maxnorm) | 7.4 | 0.036 | 16.4 | 0.035 |

Table 2: Comparison of NAG with Adaptive Gradient (AG) across several data sets. For each data set, the first line in the table contains the results using the original data, and the second line contains the results using a max-norm pre-normalized version of the original data. For both algorithms, $\eta^*$ is the optimal in-hindsight learning rate for minimizing progressive validation loss (empirically determined). Significance (bolding) was determined using a relative entropy Chernoff bound with a 0.1 probability of bound failure.

| Dataset | sNAG | | AG | |
|---|---|---|---|---|
| | $\eta^*$ | Loss | $\eta^*$ | Loss |
| Bank | 0.3 | **0.098** | $5.5 * 10^{-5}$ | 0.109 |
| (Sq norm) | 0.3 | 0.098 | 0.033 | 0.097 |
| Census | 0.11 | **0.050** | $1.2 * 10^{-6}$ | 0.054 |
| (Sq norm) | 0.11 | 0.050 | $1.6 * 10^{-3}$ | 0.048 |
| Covertype | 2.2 | **0.28** | $5.6 * 10^{-7}$ | 0.32 |
| (Sq norm) | 2.7 | 0.28 | 0.04 | 0.28 |
| CT Slice | 2.7 | **0.0019** | 0.022 | 0.0023 |
| (Sq norm) | 2.7 | 0.0019 | 0.0067 | 0.0019 |
| MSD | 7.4 | 0.0119 | $5.5 * 10^{-7}$ | 0.0130 |
| (Sq norm) | 7.4 | 0.0118 | 0.05 | 0.0120 |
| Shuttle | 11 | **0.026** | $7.5 * 10^{-4}$ | 0.040 |
| (Sq norm) | 9 | 0.026 | 0.818 | 0.026 |

Table 3: (Online Mean Square Normalized) Comparison of NAG with Adaptive Gradient (AG) across several data sets. For each data set, the first line in the table contains the results using the original data, and the second line contains the results using a squared-norm pre-normalized version of the original data. For both algorithms, $\eta^*$ is the optimal in-hindsight learning rate for minimizing progressive validation loss (empirically determined). Significance (bolding) was determined using a relative entropy Chernoff bound with a 0.1 probability of bound failure.

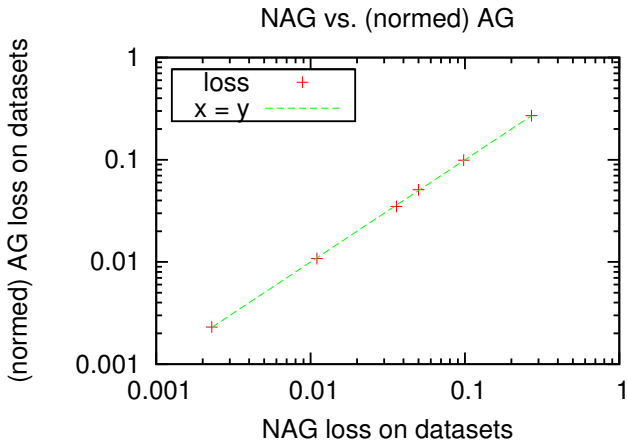

Figure 2: A comparison of performance of NAG and pre-normed AG. The results are identical, indicating that NAG effectively obsoletes pre-normalization.

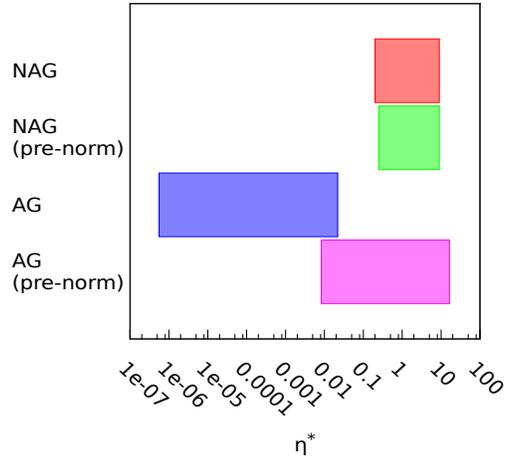

Figure 3: Each color represents the range of optimal in-hindsight learning rates $\eta^*$ across the datasets for the different learning algorithms. NAG exhibits a much smaller range of optimal values even when the data sets are pre-normalized, easing the problem of hyperparameter selection.

Some trends are evident from table 2. First, the normalized learning rule (as expected) has highest impact when the individual feature scales are highly disparate, such as data assembled from heterogeneous sensors or measurements. For instance, the CT slice data set exhibits essentially no difference; although CT slice contains physical measurements, they are histograms of raw readings from a single device, so the differences between feature ranges is modest (see table 1). Conversely the Covertype dataset shows a 5% decrease in multiclass 0-1 loss over the course of training. Covertype contains some measurements in units of meters and others in degrees, several "hillshade index" values that range from 0 to 255, and categorical variables.

The second trend evident from table 2 and reproduced in figure 3 is that the optimal learning rate is both closer to 1 in absolute terms, and varies less in relative terms, between data sets. This substantially eases the burden of tuning the learning rate for high performance. For example, a randomized search [Bergstra and Bengio, 2012] is much easier to conduct given that the optimal value is extremely likely to be within $[0.01, 10]$ independent of the data set.

The last trend evident from table 2 is the typical indifference of the normalized learning rate to pre-normalization, specifically the optimal learning rate and resulting progressive validation loss. In addition pre-normalization effectively eliminates the difference between the normalized and Adagrad updates, indicating that the online algorithm achieves results similar to the transductive algorithm for max norm.

For comprehensiveness, we also compared sNAG with a squared norm pre-normalizer, and found the story much the same in table 3. In particular sNAG dominated AG on most of the datasets and performed similarly to AG when the data was pre-normalized with a squared norm (standard deviation). It is also interesting to observe that sNAG performs slightly better than NAG on a few datasets, agreeing with our intuition that it should be more robust to outliers. Empirically, sNAG appears somewhat more robust than NAG at the cost of somewhat more computation.

## 7 Summary

We evaluated performance of Normalized Adaptive Gradient (NAG) on the most difficult unnormalized public datasets available and found that it provided performance similar to Adaptive Gradient (AG) applied to pre-normalized datasets while simultaneously collapsing the range hyperparameter search required to achieve good performance. Empirically, this makes NAG a capable and reliable learning algorithm.

We also defined a scaling adversary and proved that our algorithm is robust and efficient against this scaling adversary unlike other online learning algorithms.

### Acknowledgements

We would like to thank Miroslav Dudik for helpful discussions.